\documentclass{article}
\usepackage{ijcai16}

\usepackage{times}
\usepackage{color}
\usepackage{graphicx}
\usepackage{amsmath}
\usepackage{amsthm}
\usepackage{amssymb}
\usepackage{algorithm}
\usepackage{algorithmic}
\usepackage{subfigure}
\usepackage{url}
\usepackage{bm}
\usepackage{bbm}
\usepackage{multirow}
 \usepackage{url}
 \newtheorem{thm}{Theorem}[section]
 
 \usepackage{times}
\title{Learning A Deep $\ell_\infty$ Encoder for Hashing}

\author{Zhangyang Wang\dag, Yingzhen Yang\dag, Shiyu Chang\dag, Qing Ling\ddag, and Thomas S. Huang\dag
 \\
\dag Beckman Institute, University of Illinois at Urbana-Champaign, Urbana, IL 61801, USA\\
\ddag Department of Automation, University of Science and Technology of China, Hefei, 230027, China
}

\begin{document}

\maketitle

\begin{abstract}
We investigate the $\ell_\infty$-constrained representation which demonstrates robustness to quantization errors, utilizing the tool of deep learning. Based on the Alternating Direction Method of Multipliers (ADMM), we formulate the original convex minimization problem as a feed-forward neural network, named \textit{Deep $\ell_\infty$ Encoder}, by introducing the novel Bounded Linear Unit (BLU) neuron and modeling the Lagrange multipliers as network biases. Such a structural prior acts as an effective network regularization, and facilitates the model initialization. We then investigate the effective use of the proposed model in the application of hashing, by coupling the proposed encoders under a supervised pairwise loss, to develop a \textit{Deep Siamese $\ell_\infty$ Network}, which can be optimized from end to end. Extensive experiments demonstrate the impressive performances of the proposed model. We also provide an in-depth analysis of its behaviors against the competitors.  
\end{abstract}

\section{Introduction}

\subsection{Problem Definition and Background}

While $\ell_0$ and $\ell_1$ regularizations have been well-known and successfully applied in sparse signal approximations, it has been less explored to utilize the $\ell_\infty$ norm to regularize signal representations. In this paper, we are particularly interested in the following $\ell_\infty$-constrained least squares problem:
\begin{equation}
\begin{array}{l}\label{form}
\min_{x} ||\mathbf{D} x - y||_2^2 \quad s.t. \quad ||x||_\infty \le \lambda,
\end{array}
\end{equation}
where $y \in R^{n \times 1}$ denotes the input signal, $\mathbf{D} \in R^{n \times N}$ the (overcomplete) the basis (often called frame or dictionary) with $N < n$, and $x \in R^{N \times 1}$ the learned representation. Further, the maximum absolute magnitude of $x$ is bounded by a positive constant $\lambda$, so that each entry of $x$ has the smallest dynamic range \cite{UP}. As a result, The model (\ref{form}) tends to spread the information of $y$ approximately evenly among the coefficients of $x$.  Thus, $x$ is called ``democratic'' \cite{Wotao} or ``anti-sparse'' \cite{fuchs2011spread}, as all of its entries are of approximately the same importance.

In practice, $x$ usually has most entries reaching the same absolute maximum magnitude \cite{Wotao}, therefore resembling to an antipodal signal in an $N$-dimensional Hamming space. Furthermore, the solution $x$ to (\ref{form}) withstands errors in a very powerful way: the representation error gets bounded by the average, rather than the sum, of the errors in the coefficients. These errors may be of arbitrary nature, including distortion (e.g., quantization) and losses (e.g., transmission failure). This property was quantitatively established in Section II.C of \cite{UP}:
\begin{thm} \label{E}
Assume $||x||_2 < 1$ without loss of generality, and each coefficient of $x$ is quantized separately by performing a uniform scalar quantization of the dynamic range [$- \lambda, \lambda$] with $L$ levels. The overall quantization error of $x$ from (\ref{form}) is bounded by $\frac{\lambda \sqrt{N}}{L}$. In comparision, a least squares solution $x_{LS}$, by minimizing $||\mathbf{D} x_{LS} - y||_2^2$ without any constraint, would only give the bound $\frac{\sqrt{n}}{L}$.
\end{thm}
In the case of $N << n$, the above will yield great robustness for the solution to (\ref{form}) with respect to noise, in particular quantization errors. Also note that its error bound will not grow with the input dimensionality $n$, a highly desirable stability property for high-dimensional data. Therefore, (\ref{form}) appears to be favorable for the applications such as vector quantization, hashing and approximate nearest neighbor search. 

% our contribution
In this paper, we investigate (\ref{form}) in the context of deep learning. Based on the Alternating Direction Methods of Multipliers (\textbf{ADMM}) algorithm, we formulate (\ref{form}) as a feed-forward neural network \cite{LISTA}, called \textbf{Deep $\ell_\infty$ Encoder}, by introducing the novel Bounded Linear Unit (\textbf{BLU}) neuron and modeling the Lagrange multipliers as network biases. The major technical merit to be presented, is how a specific optimization model (\ref{form}) could be translated to designing a task-specific deep model, which displays the desired quantization-robust  property. We then study its application in hashing, by developing a \textbf{Deep Siamese $\ell_\infty$ Network} that couples the proposed encoders under a supervised pairwise loss, which could be optimized from end to end. Impressive performances are observed in our experiments.

%We would also like to emphasize that this is not an ``application-driven'' paper, i.e., one dedicated to improving hashing performance. The major technical merit to be presented, is how a specific optimization model (\ref{form}) could be translated to designing a task-specific deep model, which displays the desired quantization-robust  property.  

\subsection{Related Work}

% algorithm: prime dual, active set
Similar to the case of $\ell_0$/$\ell_1$ sparse approximation problems, solving (\ref{form}) and its variants (e.g., \cite{Wotao}) relies on iterative solutions. \cite{stark1995bounded} proposed an active set strategy similar to that of \cite{lawson1974solving}. In \cite{adlers1998sparse}, the authors investigated a primal-dual path-following interior-point method. Albeit effective, the iterative approximation algorithms suffer from their inherently sequential structures, as well as the data-dependent complexity and latency, which often constitute a major bottleneck in the computational efficiency. In addition, the joint optimization of the (unsupervised) feature learning and the supervised steps has to rely on solving complex bi-level optimization problems \cite{IJCAI}. Further, to effectively represent datasets of growing sizes, a larger dictionaries $\mathbf{D}$ is usually in need. Since the inference complexity of those iterative algorithms increases more than linearly with respect to the dictionary size \cite{bertsekas1999nonlinear}, their scalability turns out to be limited. Last but not least, while the hyperparameter $\lambda$ sometimes has physical interpretations, e.g., for signal quantization and compression, it remains unclear how to be set or adjusted for many application cases.  

Deep learning has recently attracted great attentions \cite{ImageNet}. The advantages of deep learning lie in its composition of multiple non-linear transformations to yield more abstract and descriptive embedding representations. The feed-forward networks could be naturally tuned jointly with task-driven loss functions \cite{AAAI16}. With the aid of gradient descent, it also scales linearly in time and space with the number of train samples. 

There has been a blooming interest in bridging ``shallow'' optimization and deep learning models. In \cite{LISTA}, a feed-forward neural network, named LISTA, was proposed to efficiently approximate the sparse codes, whose hyperparameters were learned from general regression. In \cite{sprechmann2013supervised}, the authors leveraged a similar idea on fast trainable regressors and constructed feed-forward network approximations of the learned sparse models. It was later extended in \cite{PAMI2015} to develop a principled process of learned deterministic fixed-complexity pursuits, for sparse and low rank models. Lately, \cite{AAAI16} proposed Deep $\ell_0$ Encoders, to model $\ell_0$ sparse approximation as feed-forward neural networks. The authors extended the strategy to the graph-regularized $\ell_1$ approximation in \cite{SDM}, and the dual sparsity model in \cite{D3}. Despite the above progress, up to our best knowledge, few efforts have been made beyond sparse approximation (e.g., $\ell_0$/$\ell_1$) models.

\section{An ADMM Algorithm}

ADMM has been popular for its remarkable effectiveness in minimizing objectives with linearly separable structures \cite{bertsekas1999nonlinear}. We first introduce an auxiliary variable $z \in R^{N \times 1}$, and rewrite (\ref{form}) as:
\begin{equation}
\begin{array}{l}\label{unconstrain}
\min_{\mathbf{x, z}}  \frac{1}{2}||\mathbf{D} x - y||_2^2 \quad s.t. \quad ||z||_\infty \le \lambda, \quad z - x = 0.
\end{array}
\end{equation}
The augmented Lagrangian function of (\ref{unconstrain}) is:
\begin{equation}
\begin{array}{l}\label{aug}
 \frac{1}{2}||\mathbf{D} x - y||_2^2 + p^T (z - x ) + \frac{\beta}{2}|| z -x ||^2_2 + \Phi_\lambda(z).
\end{array}
\end{equation}
Here $p \in R^{N \times 1}$ is the Lagrange multiplier attached to the equality constraint, $\beta$ is a positive constant (with a default value 0.6), and $\Phi_\lambda(z)$ is the indicator function which goes to infinity when $||z||_\infty > \lambda$ and 0 otherwise. ADMM minimizes (\ref{aug}) with respect to $x$ and $z$ in an alternating direction manner, and updates $p$ accordingly. It guarantees global convergence to the optimal solution to (\ref{form}).
Starting from any initialization points of $x$, $z$, and $p$, ADMM iteratively solves ($t$ = 0, 1, 2... denotes the iteration number):
\begin{equation}
\begin{array}{l}\label{solvex}
\text{\textit{x update:}} \quad \min_{x_{t+1}}  \frac{1}{2}||\mathbf{D} x - y||_2^2 - p_t^T x + \frac{\beta}{2}|| z_t -x ||^2_2,
\end{array}
\end{equation}
\begin{equation}
\begin{array}{l}\label{solvez}
\text{\textit{z update:}} \quad \min_{z_{t+1}} \frac{\beta}{2}|| z - (x_{t+1} - \frac{p_t}{\beta}) ||^2_2 + \Phi_\lambda(z),
\end{array}
\end{equation}
\begin{equation}
\begin{array}{l}\label{solvep}
\text{\textit{p update:}} \quad p_{t+1} = p_t + \beta ( z_{t+1} -x_{t+1}).
\end{array}
\end{equation}
Furthermore, both (\ref{solvex}) and (\ref{solvez}) enjoy closed-form solutions: 
\begin{equation}
\begin{array}{l}\label{rulex}
x_{t+1} = (\mathbf{D}^T \mathbf{D} + \beta \mathbf{I})^{-1} (\mathbf{D}^T y + \beta z_t + p_t),
\end{array}
\end{equation}
\begin{equation}
\begin{array}{l}\label{rulez}
z_{t+1} = \min(\max(x_{t+1} - \frac{p_t}{\beta}, - \lambda), \lambda).
\end{array}
\end{equation}
%This algorithm guarantees global convergence to the optimal solution to \ref{form}.

%\cite{bertsekas1999nonlinear}

%The algorithm belongs to a form of the primal-dual hybrid gradient algorithm in\cite{goldstein2013adaptive}.

The above algorithm could be categorized to the primal-dual scheme. However, discussing the ADMM algorithm in more details is beyond the focus of this paper. Instead, the purpose of deriving (\ref{unconstrain})-(\ref{rulez}) is to preparing us for the design of the task-specific deep architecture, as presented below.

\section{Deep $\ell_\infty$ Encoder}

We first substitute (\ref{rulex}) into (\ref{rulez}), in order to derive an update form explicitly dependent on only $z$ and $p$:
\begin{equation}
\begin{array}{l}\label{iter}
z_{t+1} = B_\lambda((\mathbf{D}^T \mathbf{D} + \beta \mathbf{I})^{-1} (\mathbf{D}^T y + \beta z_t + p_t) - \frac{p_t}{\beta}),
\end{array}
\end{equation}
where $B_{\lambda}$ is defined as a box-constrained element-wise operator ($u$ denotes a vector and $u_i$ is its $i$-th element):
\begin{equation}
\label{box}
[B_\lambda (u)]_i = \min(\max(u_i, - \lambda), \lambda).
\end{equation}

\begin{figure}[htbp]
\centering
\begin{minipage}{0.30\textwidth}
\includegraphics[width=\textwidth]{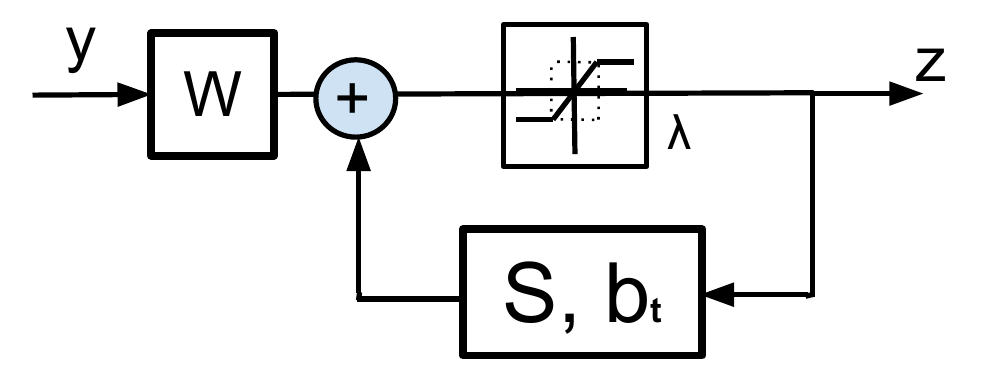}
\end{minipage}
\caption{The block diagram of solving solving (\ref{form}). }.
\label{figISTA}
\end{figure}

\begin{figure*}[tbp]
\centering
\begin{minipage}{0.90\textwidth}
\includegraphics[width=\textwidth]{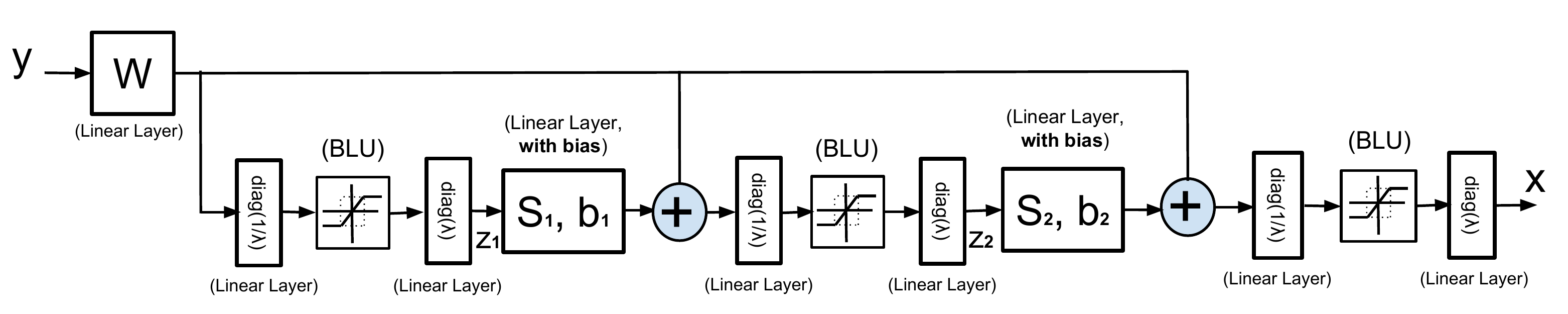}
\end{minipage}
\caption{Deep $\ell_\infty$ Encoder, with two time-unfolded stages.}
\label{figLISTA}
\end{figure*}

\begin{figure}[tbp]
\centering
\begin{minipage}{0.48\textwidth}
\includegraphics[width=\textwidth]{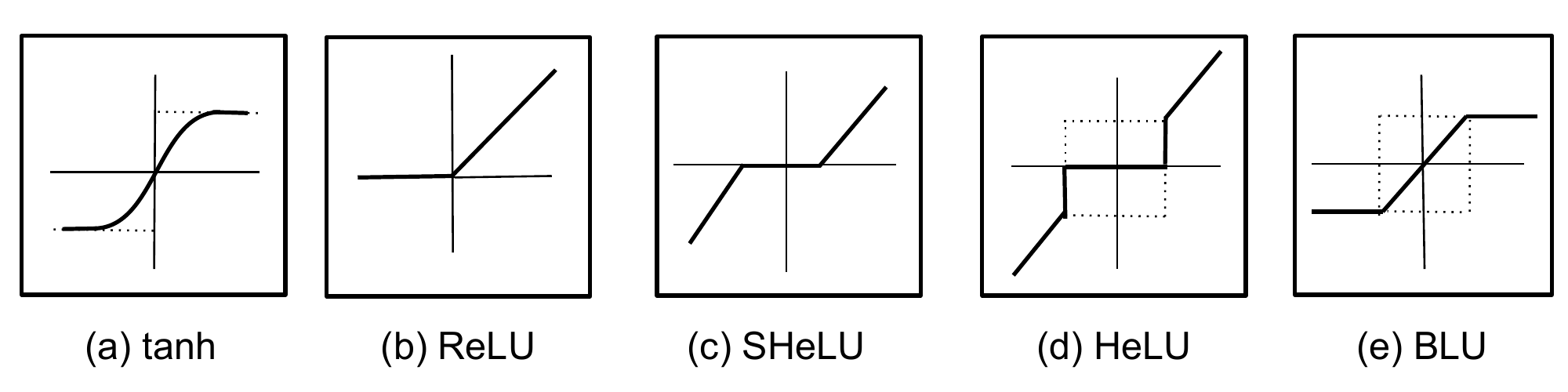}
\end{minipage}
\caption{A comparison among existing neurons and BLU.}
\label{neuron}
\end{figure}

Eqn. (\ref{iter}) could be alternatively rewritten as:
\begin{equation}
\begin{array}{l}\label{recur}
\mathbf{z}_{t+1} = B_\lambda(\mathbf{W} y + \mathbf{S} z_t + b_t), $ $\text{where:} \\
\mathbf{W} = (\mathbf{D}^T \mathbf{D} + \beta \mathbf{I})^{-1} \mathbf{D}^T, \mathbf{S} = \beta (\mathbf{D}^T \mathbf{D} + \beta \mathbf{I})^{-1}, 
\\b_t = [(\mathbf{D}^T \mathbf{D} + \beta \mathbf{I})^{-1} - \frac{1}{\beta}\mathbf{I}] p_t,
\end{array}
\end{equation}
and expressed as the block diagram in Fig. \ref{figISTA}, which outlines a recurrent structure of solving (\ref{form}). Note that in (\ref{recur}), while $\mathbf{W}$ and $\mathbf{S}$ are pre-computed hyperparamters shared across iterations, $b_t$ remains to be a variable dependent on $p_t$, and has to be updated throughout iterations too ($b_t$'s update block is omitted in Fig. \ref{figISTA}).

By time-unfolding and truncating Fig. \ref{figISTA} to a fixed number of $K$ iterations ($K$ = 2 by default)\footnote{We test larger $K$ values (3 or 4). In several cases they do bring performance improvements, but add complexity too.}, we obtain a feed-forward network structure in Fig. \ref{figLISTA}, named \textbf{Deep $\ell_\infty$ Encoder}. Since the threshold $\lambda$ is less straightforward to update, we repeat the same trick in \cite{AAAI16} to rewrite (\ref{box}) as: $[B_\lambda (u)]_i = \lambda_i B_1(u_i/ \lambda_i)$. The original operator is thus decomposed into two linear diagonal scaling layers, plus a unit-threshold neuron, the latter of which is called a Bounded Linear Unit (\textbf{BLU}) by us.  All the hyperparameters $\mathbf{W}$,  $\mathbf{S}_k$ and $b_k$ ($k$ = 1, 2), as well as $\lambda$, are all to be learnt from data by back-propogation. Although the equations in (\ref{recur}) do not directly apply to solving the deep $\ell_\infty$ encoder, they can serve as high-quality initializations. 

\begin{figure}[tbp]
\centering
\begin{minipage}{0.40\textwidth}
\includegraphics[width=\textwidth]{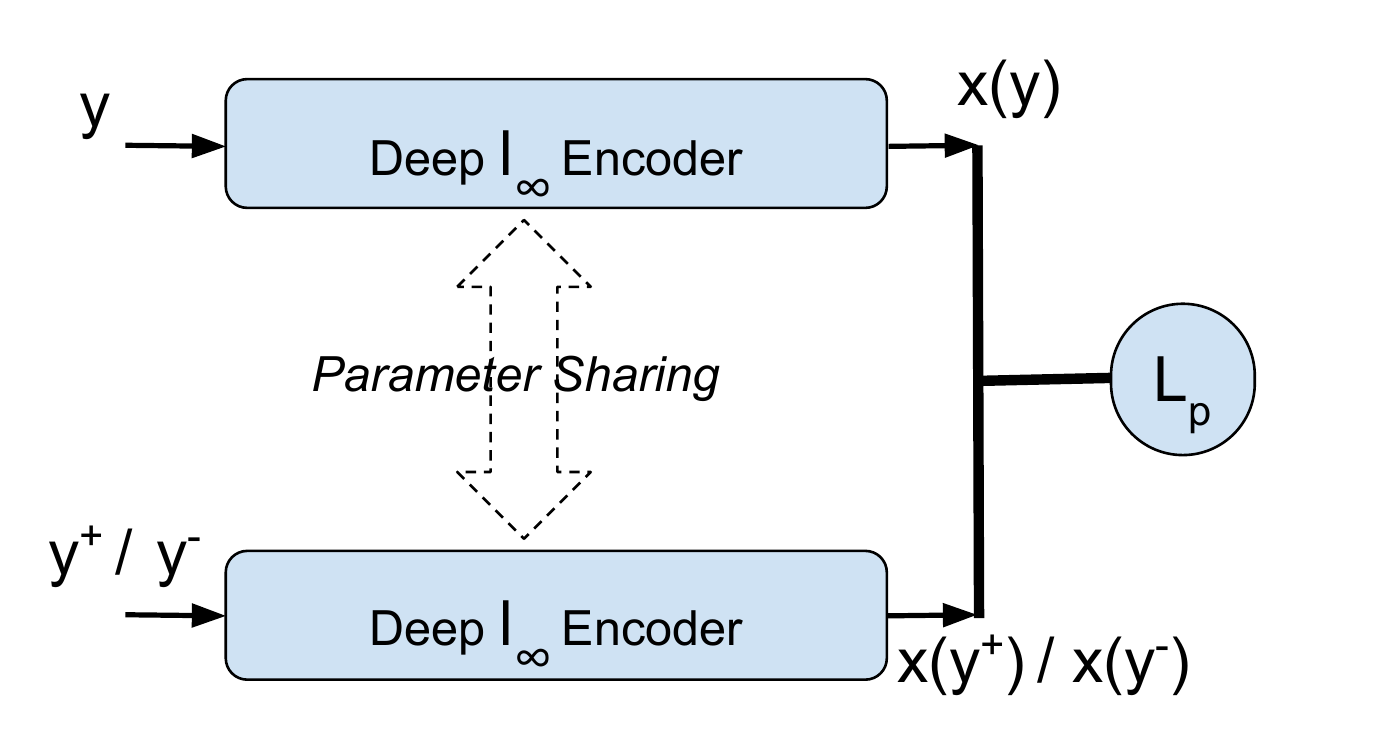}
\end{minipage}
\caption{Deep $\ell_\infty$ Siamese Network, by coupling two parameter-sharing encoders, followed by a pairwise loss (\ref{ploss}).}
\label{siamese}
\end{figure}

It is crucial to notice the modeling of the Lagrange multipliers $p_t$ as the biases, and to incorporate its updates into network learning. That provides important clues on how to relate deep networks to a larger class of optimization models, whose solutions rely on dual domain methods.

%\noindent \textbf
\noindent \textbf{Comparing BLU with existing neurons} As shown in Fig. \ref{neuron} (e), BLU tends to suppress large entries while not penalizing small ones, resulting in dense, nearly antipodal representations. A first look at the plot of BLU easily reminds the tanh neuron (Fig. \ref{neuron} (a)). In fact, with the its output range $[-1,1]$ and a slope of 1 at the origin, tanh could be viewed as a smoothened differentiable approximation of BLU. 

We further compare BLU with other popular and recently proposed neurons: Rectifier Linear Unit (ReLU) \cite{ImageNet}, Soft-tHresholding Linear Unit (SHeLU) \cite{SDM}, and Hard thrEsholding Linear Unit (HELU) \cite{AAAI16}, as depicted in Fig. \ref{neuron} (b)-(d), respectively. Contrary to BLU and tanh, they all introduce sparsity in the outputs, and thus prove successful and outperform tanh in classification and recognition tasks. Interestingly, HELU seems exactly the rival against BLU, as it does not penalize large entries but suppresses small ones down to zero.

%RELU could be viewed as a translated variant of SHeLU, that is further enforced with non-negative constraints. 

% sparse versus dense

\section{Deep $\ell_\infty$ Siamese Network for Hashing}

Rather than solving (\ref{form}) first and then training the encoder as general regression, as  \cite{LISTA} did, we instead concatenate encoder(s) with a task-driven loss, and optimize the pipeline \textbf{from end to end}. In this paper, we focus on discussing its application in hashing, although the proposed model is not limited to one specific application.

% review problem; prior art (shallow)

\noindent \textbf{Background} With the ever-growing large-scale image data on the Web, much attention has been devoted to nearest neighbor search via hashing methods \cite{LSH}. For big data applications, compact bitwise representations improve the efficiency in both storage and search speed. The state-of-the-art approach, learning-based hashing, learns similarity-preserving hash functions to encode input data into binary codes. Furthermore, while earlier methods, such as linear search hashing (LSH) \cite{LSH}, iterative quantization (ITQ) \cite{ITQ} and spectral hashing (SH) \cite{SH}, do not refer to any supervised information, it has been lately discovered that involving the data similarities/dissimilarities in training benefits the performance \cite{kulis2009learning,KSH}.

\noindent \textbf{Prior Work} Traditional hashing pipelines first represent each input image as a (hand-crafted) visual descriptor, followed by separate projection and quantization steps to encode it into a binary code. \cite{NNHash} first applied the \textit{siamese network} \cite{hadsell2006dimensionality} architecture to hashing, which fed two input patterns into two parameter-sharing ``encoder'' columns and minimized a pairwise-similarity/dissimilarity loss function between their outputs, using pairwise labels. The authors further enforced the sparsity prior on the hash codes in \cite{masci2013sparse}, by substituting a pair of LISTA-type encoders \cite{LISTA} for the pair of generic feed-forward encoders in \cite{NNHash} \cite{xia2014supervised,li2015feature} utilized tailored convolution networks with the aid of pairwise labels. \cite{lai2015simultaneous} further introduced a triplet loss with a divide-and-encode strategy applied to reduce the hash code redundancy. Note that for the final training step of quantization, \cite{masci2013sparse} relied on an extra hidden layer of tanh neurons to approximate binary codes, while \cite{lai2015simultaneous} exploited a piece-wise linear and discontinuous threshold function.

\noindent \textbf{Our Approach} In view of its robustness to quantization noise, as well as BLU's property as a natural binarization approximation, we construct a siamese network as in \cite{NNHash}, and adopt a pair of parameter-sharing deep $\ell_\infty$ encoders as the two columns. The resulting architecture, named the \textbf{Deep $\ell_\infty$ Siamese Network}, is illustrated in Fig. \ref{siamese}. Assume $y$ and $y^+$ make a similar pair while $y$ and $y^-$ make a dissimilar pair, and further denote $x(y)$ the output representation by inputting $y$. The two coupled encoders are then optimized under the following pairwise loss (the constant $m$ represents the margin between dissimilar pairs):
\begin{equation}
\begin{array}{l}\label{ploss}
L_p := \frac{1}{2}||x(y) - x(y^+)||^2 - \\ \frac{1}{2}(\max(0, m - ||x(y) - x(y^-)||))^2.
\end{array}
\end{equation}
The representation is learned to make similar pairs as close as possible and dissimilar pairs at least at distance $m$. In this paper, we follow \cite{NNHash} to use a default $m$ = 5 for all experiments. 

Once a deep $\ell_\infty$ siamese network is learned, we apply its encoder part (i.e., a deep $\ell_\infty$ encoder) to a new input. The computation is extremely efficient, involving only a few matrix multiplications and element-wise thresholding operations, with a total complexity of $O(nN + 2N^2)$. One can obtain a $N$-bit binary code by simply quantizing the output.

%\begin{figure}[htbp]
%\centering
%\begin{minipage}{0.30\textwidth}
%\centering \subfigure[] {
%\includegraphics[width=\textwidth]{LS.png}
%}\\
%\centering \subfigure[] {
%\includegraphics[width=\textwidth]{LShashing.png}
%}\end{minipage}
%\begin{minipage}{0.30\textwidth}
%\centering \subfigure[] {
%\includegraphics[width=\textwidth]{LS.png}
%}\\
%\centering \subfigure[] {
%\includegraphics[width=\textwidth]{LShashing.png}
%}\end{minipage}
%\begin{minipage}{0.30\textwidth}
%\centering \subfigure[] {
%\includegraphics[width=\textwidth]{LS.png}
%}\\
%\centering \subfigure[] {
%\includegraphics[width=\textwidth]{LShashing.png}
%}\end{minipage}
%\caption{outputs}
%\end{figure}

 \begin{table}[htbp]
 \begin{center}
 \caption{Comparison of NNH, SNNH, and the proposed deep $\ell_\infty$ siamese network.}
 \label{encoder}
 \begin{tabular}{|c|c|c|c|}
 \hline
$ $ & encoder  &  neuron &   structural prior   \\
$ $ & type &   type  &   on hashing codes   \\
\hline
NNH & generic & tanh & \slash  \\
\hline
SNNH & LISTA & SHeLU & sparse  \\
\hline
\multirow{2}{*}{Proposed} & \multirow{2}{*}{deep $\ell_\infty$} & \multirow{2}{*}{BLU} & nearly antipodal \\
$$ &$$ &$$ & \& quantization-robust\\
\hline
 \end{tabular}
 \end{center}
 \end{table}

\section{Experiments in Image Hashing}

\noindent \textbf{Implementation}
The proposed networks are implemented with the CUDA ConvNet package \cite{ImageNet}. We use a constant learning rate of 0.01 with no momentum, and a batch size of 128. Different from prior findings such as in \cite{AAAI16,SDM}, we discover that untying the values of $\mathbf{S}_1$, $b_1$ and $\mathbf{S}_2$, $b_2$ boosts the performance more than sharing them. It is not only because that more free parameters enable a larger learning capacity, but also due to the important fact that $p_t$ (and thus $b_k$) is in essence not shared across iterations, as in (\ref{recur}) and Fig. \ref{figLISTA}.

While many neural networks are trained well with random initializations, it has been discovered that sometimes poor initializations can still hamper the effectiveness of first-order methods \cite{sutskever2013importance}. On the other hand, It is much easier to initialize our proposed models in the right regime. We first estimate the dictionary $\mathbf{D}$ using the standard K-SVD algorithm \cite{KSVD}, and then inexactly solve (\ref{form}) for up to $K$ ($K$ = 2) iterations, via the ADMM algorithm in Section 2, with the values of Lagrange multiplier $p_t$ recorded for each iteration. Benefiting from the analytical correspondence relationships in (\ref{recur}), it is then straightforward to obtain high-quality initializations for $\mathbf{W}$, $\mathbf{S}_k$ and $b_k$ ($k$ = 1, 2). As a result, we could achieve a steadily decreasing curve of training errors, without performing common tricks such as annealing the learning rate, which are found to be indispensable if random initialization is applied.

\begin{table*}[htbp]
\vspace{0mm}
\small
\caption{Performance (\%)  of different hashing methods on the CIFAR10 dataset, with different code lengths $N$. 
\vspace{0mm}
}
\begin{center}
\begin{tabular}{cc  ccc  | c   c  c c c c c c}
&& &&&		 &   \multicolumn{3}{c}{{\bf Hamming radius} $\leq 2$} & \multicolumn{3}{c}{{\bf Hamming radius} $=0$} \\
\cline{7-12}
 \multicolumn{5}{c|}{{\bf Method}}	& {\bf mAP}  & {\bf Prec.} & {\bf Recall} & {\bf F1} & {\bf Prec.} & {\bf Recall} & {\bf F1} \\
\hline
 & & & & $N$
& & & & & & &  \\
\multirow{2}{*}{{\bf KSH}} 	
&   &&&48			
& 31.10 & 18.22 & 0.86 & 1.64 & 5.39 & 5.6$\times 10^{-2}$ &0.11 \\
&   &&&64			
& 32.49 & 10.86 & 0.13 & 0.26 & 2.49 & 9.6$\times 10^{-3}$ & 1.9$\times 10^{-2}$ \\
%&   &&&128			
%& 33.50 & 2.91 & 3.3$\times 10^{-3}$ & ?? & 0.67 & 3.4$\times 10^{-4}$ & ?? \\
%
\hline%\cline{1-12}
\multirow{2}{*}{{\bf AGH1}} 	
&   &&&48		
& 14.55 & 15.95 & 2.8$\times 10^{-2}$ & 5.6$\times 10^{-2}$ & 4.88 & 2.2$\times 10^{-3}$ & 4.4$\times 10^{-3}$ \\
&   &&&64		
& 14.22 & 6.50 & 4.1$\times 10^{-3}$ & 8.1$\times 10^{-3}$ & 3.06 & 1.2$\times 10^{-3}$ & 2.4$\times 10^{-3}$ \\
%&   &&&128			
%& 13.53 & 2.89 & 1.1$\times 10^{-3}$ & ?? & 1.58 & 4.5$\times 10^{-3}$ & ?? \\
%
\hline
\multirow{2}{*}{{\bf AGH2}} 	
&   &&&48		
& 15.34 & 17.43 & 7.1$\times 10^{-2}$ & 3.6$\times 10^{-2}$ & 5.44 & 3.5$\times 10^{-3}$ & 6.9$\times 10^{-3}$ \\
&   &&&64
& 14.99 & 7.63 & 7.2$\times 10^{-3}$ & 1.4$\times 10^{-2}$ & 3.61 & 1.4$\times 10^{-3}$ & 2.7$\times 10^{-3}$ \\
%&   &&&128			
%& 14.38 & 3.78 & 1.6$\times 10^{-3}$ & ?? & 1.43 & 3.9$\times 10^{-4}$ & ?? \\
%
\hline
\multirow{2}{*}{{\bf PSH}} 	
&   &&&48			
& 15.78 & 9.92 & 6.6$\times 10^{-3}$ & 1.3$\times 10^{-2}$ & 0.30 & 5.1$\times 10^{-5}$ & 1.0$\times 10^{-4}$ \\
&   &&&64
& 17.18 & 1.52 & 3.0$\times 10^{-4}$ & 6.1$\times 10^{-4}$ & 1.0$\times 10^{-3}$ & 1.69$\times 10^{-5}$ & 3.3$\times 10^{-5}$ \\
%&   &&&128			
%& - & - & - & ?? & - & - & ?? \\
%
\hline
\multirow{2}{*}{{\bf LH}} 	
&   &&&48	
& 13.13 & 3.0$\times 10^{-3}$ & 1.0$\times 10^{-4}$ & 5.1$\times 10^{-5}$ & 1.0$\times 10^{-3}$ & 1.7$\times 10^{-5}$ & 3.4$\times 10^{-5}$ \\
&   &&&64			
& 13.07 & 1.0$\times 10^{-3}$ & 1.7$\times 10^{-5}$ & 3.3$\times 10^{-5}$ & 0.00 & 0.00 & 0.00 \\
%&   &&&128			
%& - & - & - & ?? & - & - & ?? \\
%
\hline
%& & & & $m$ & $M$
%& & & & & &   \\
\multirow{2}{*}{{\bf NNH}} 	
&&   &&48%& 7	
&  31.21 &  34.87 & 1.81 & 3.44 & 10.02 & 9.4$\times 10^{-2}$ & 0.19 \\
%NNhash L2 (64dim) m=7	&  &  &  &  &  \\
%
%
&   &&&64
& 35.24 & 23.23 & 0.29 & 0.57 & 5.89 & 1.4$\times 10^{-2}$ & 2.8$\times 10^{-2}$ \\
%&   &&&128			
%& 37.89 & 1.39 & 2.9$\times 10^{-3}$ & ?? & 5.38 & 2.9$\times 10^{-2}$ & ?? \\
%
%
\hline
 \multirow{2}{*}{{\bf SNNH}} 	
&&   &&48%& 7	
& 26.67 & 32.03 & 12.10 & 17.56 & 19.95 & \textbf{0.96} & \textbf{1.83} \\
%NNhash L2 (64dim) m=7	&  &  &  &  &  \\
%
%
&   &&&64
& 27.25 & 30.01 & \textbf{36.68} & 33.01 & 30.25 & \textbf{9.8} & \textbf{14.90} \\
%&   &&&128			
%& 37.89 & 1.39 & 2.9$\times 10^{-3}$ & ?? & 5.38 & 2.9$\times 10^{-2}$ & ?? \\
%
%
\hline
\multirow{2}{*}{{\bf Proposed}} 	
&&   &&48%& 7	
& \textbf{31.48} & \textbf{36.89} & \textbf{12.47} & \textbf{18.41} & \textbf{24.82} & 0.94 &1.82 \\
%NNhash L2 (64dim) m=7	&  &  &  &  &  \\
%
%
&   &&&64
& \textbf{36.76} & \textbf{38.67} & 30.28 & \textbf{33.96} & \textbf{33.30} & 8.9 & 14.05 \\
%&   &&&128			
%& 37.89 & 1.39 & 2.9$\times 10^{-3}$ & ?? & 5.38 & 2.9$\times 10^{-2}$ & ?? \\
%
%
\hline
\end{tabular}
\end{center}
\label{cifar}\vspace{0mm}
\end{table*}%

\begin{figure}[htbp]
\centering \subfigure[NNH representation] {
\includegraphics[width=0.225\textwidth]{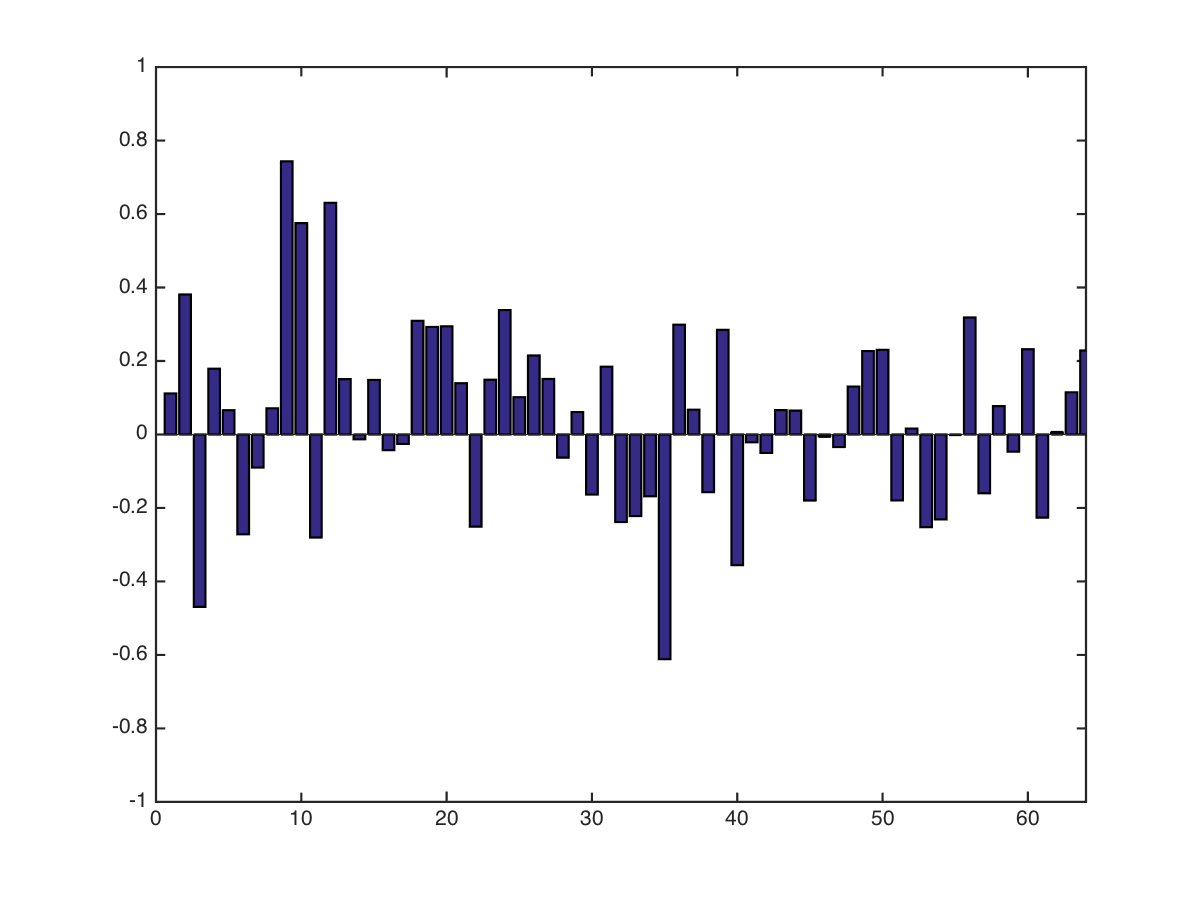}
}
\centering \subfigure[NNH binary hashing code] {
\includegraphics[width=0.23\textwidth]{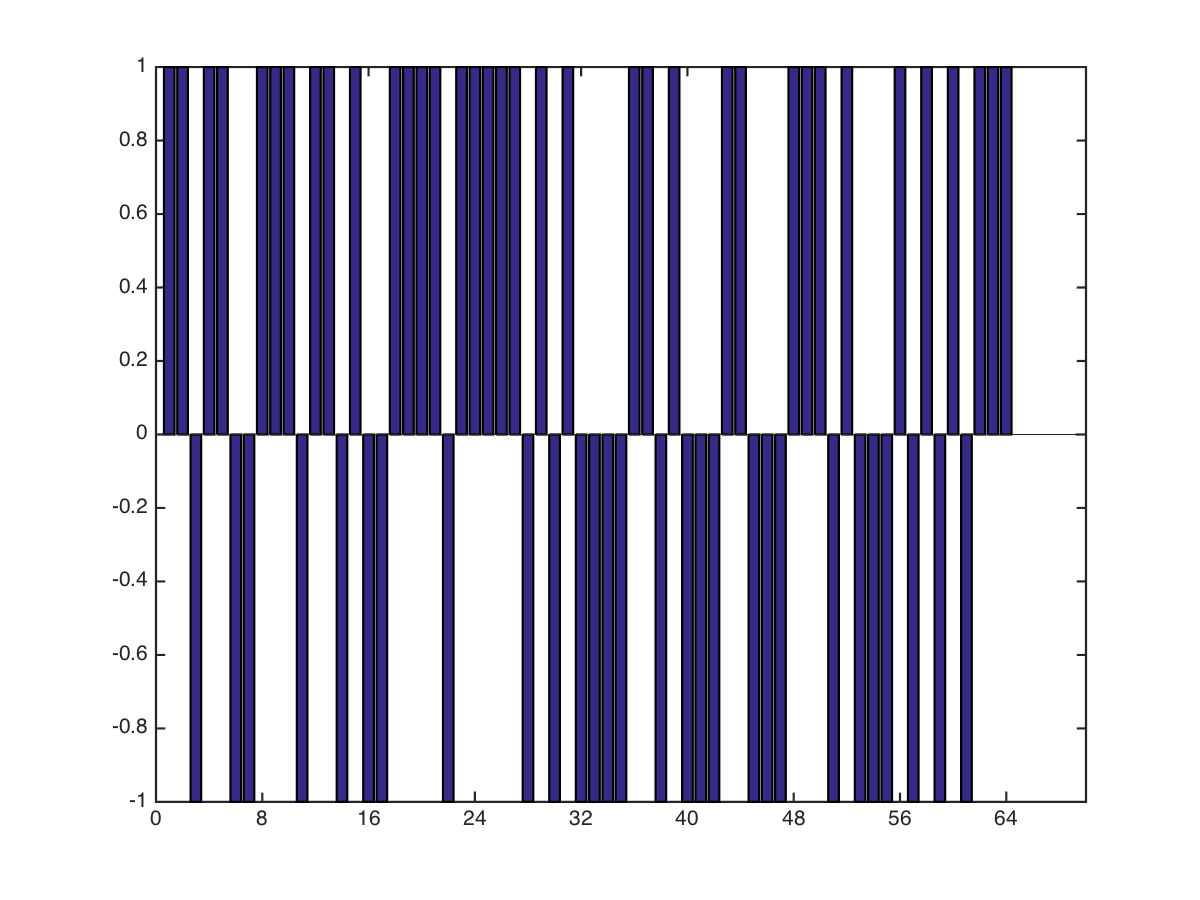}
}\\
\centering \subfigure[SNNH representation] {
\includegraphics[width=0.225\textwidth]{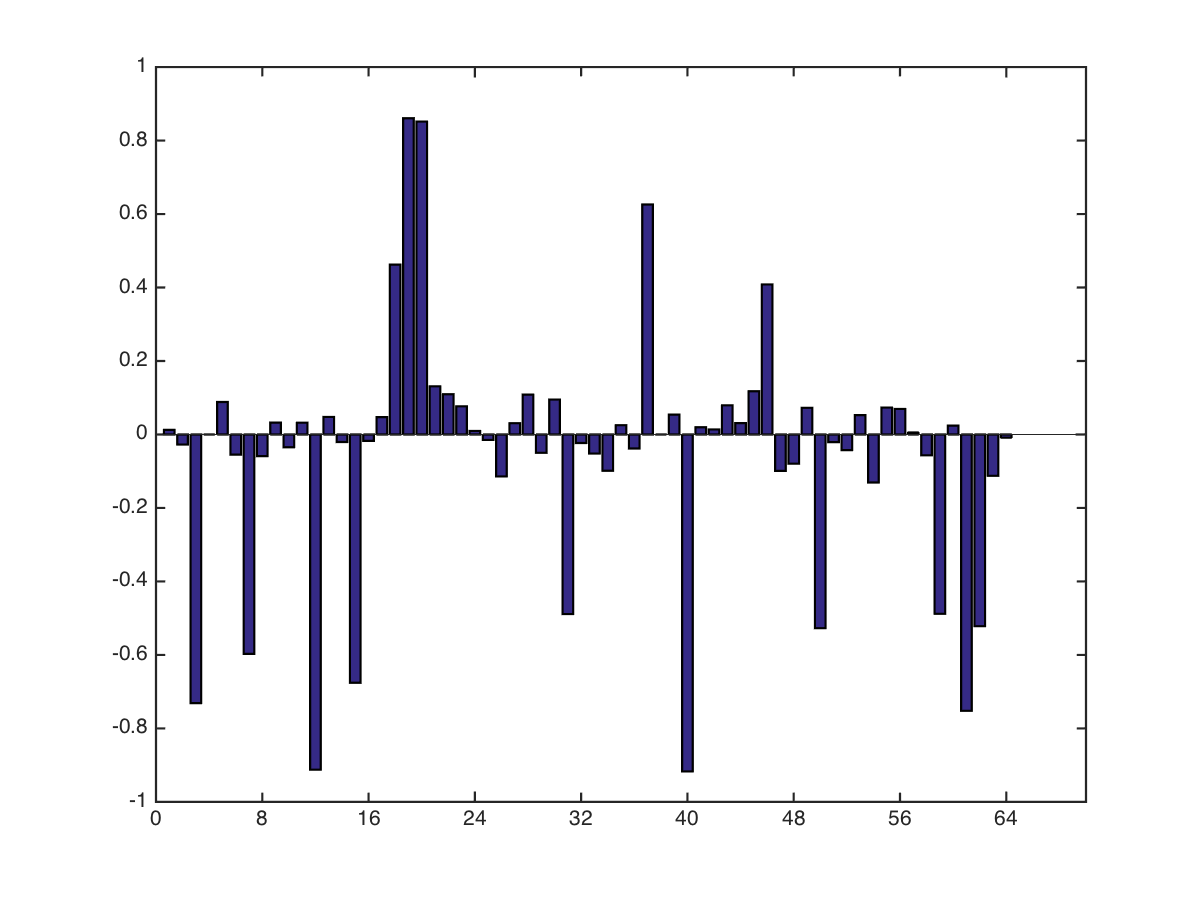}
}
\centering \subfigure[SNNH binary hashing code] {
\includegraphics[width=0.23\textwidth]{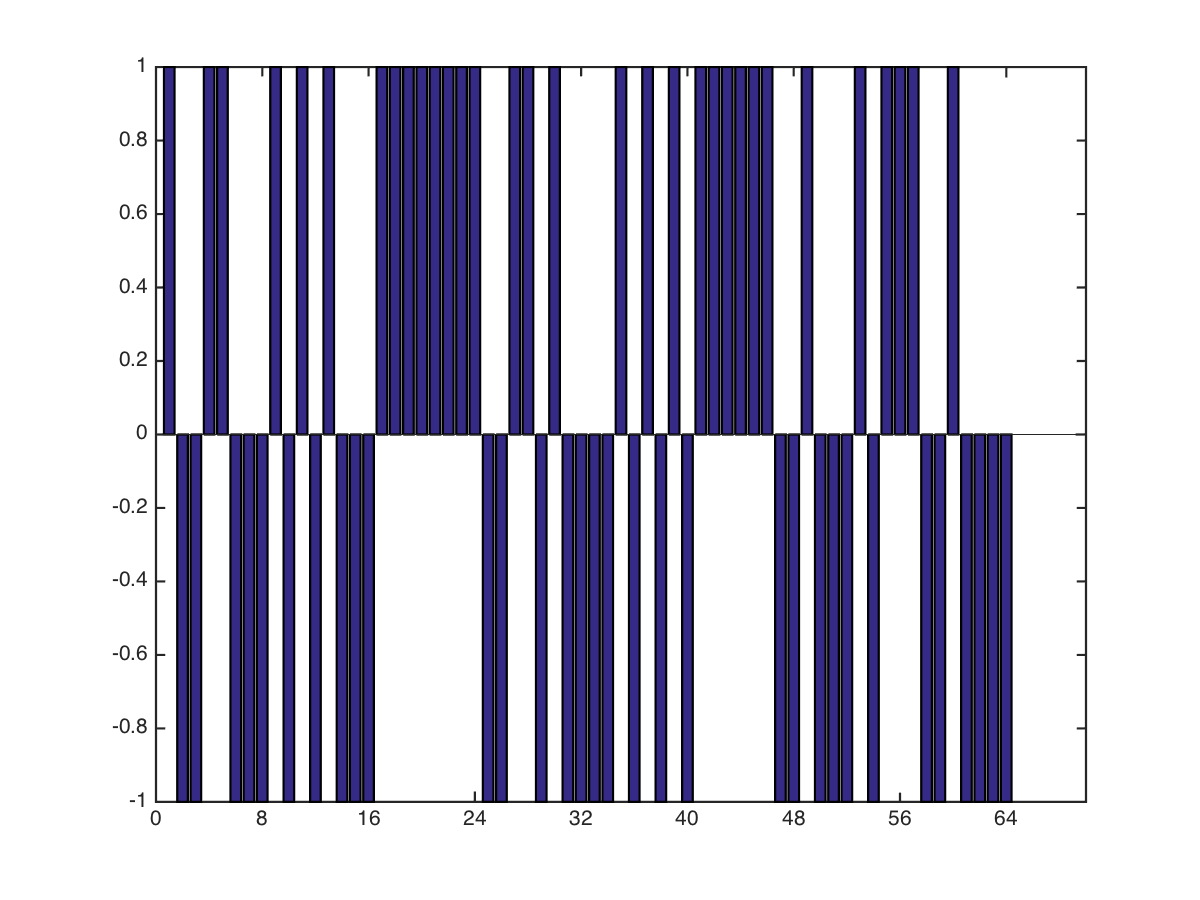}
}\\
\centering \subfigure[Deep $\ell_\infty$ representation] {
\includegraphics[width=0.225\textwidth]{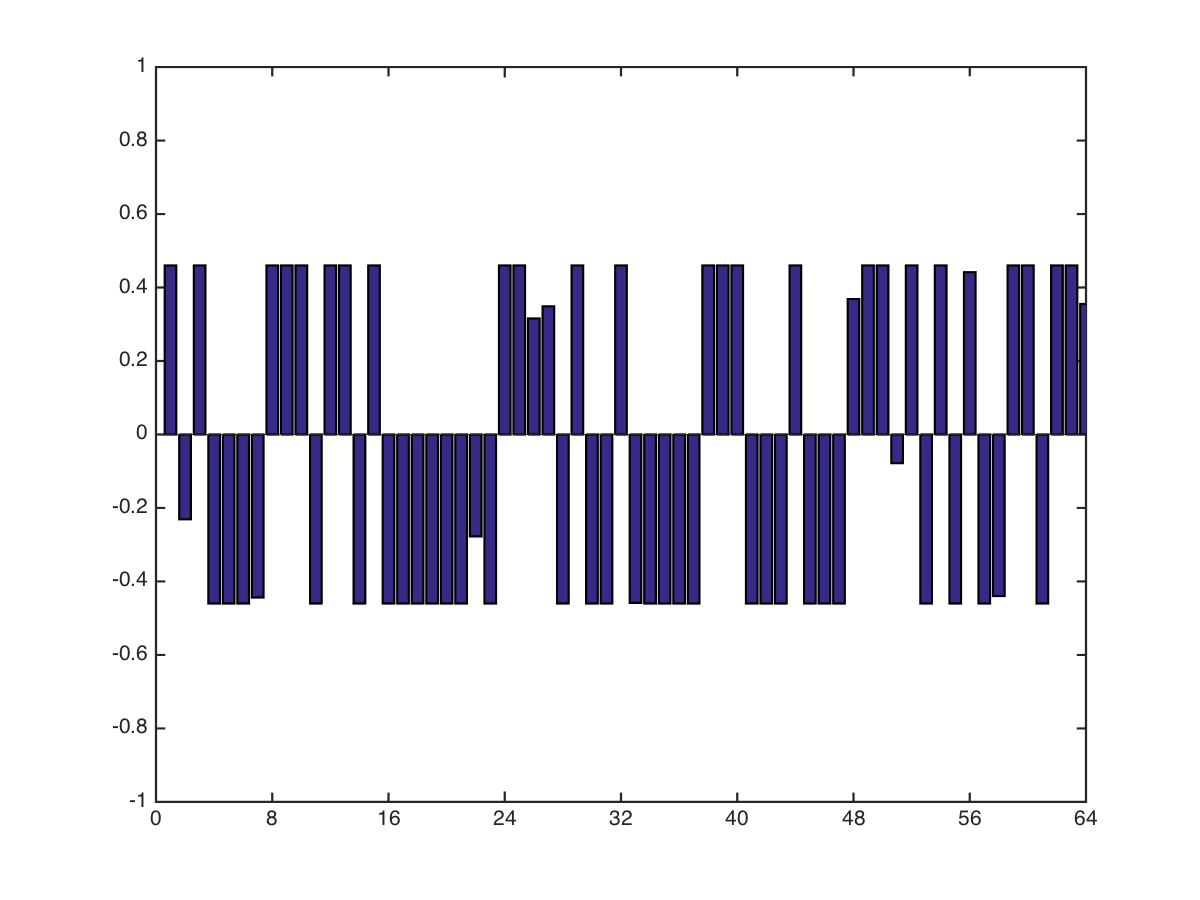}
}
\centering \subfigure[Deep $\ell_\infty$ binary hashing code] {
\includegraphics[width=0.23\textwidth]{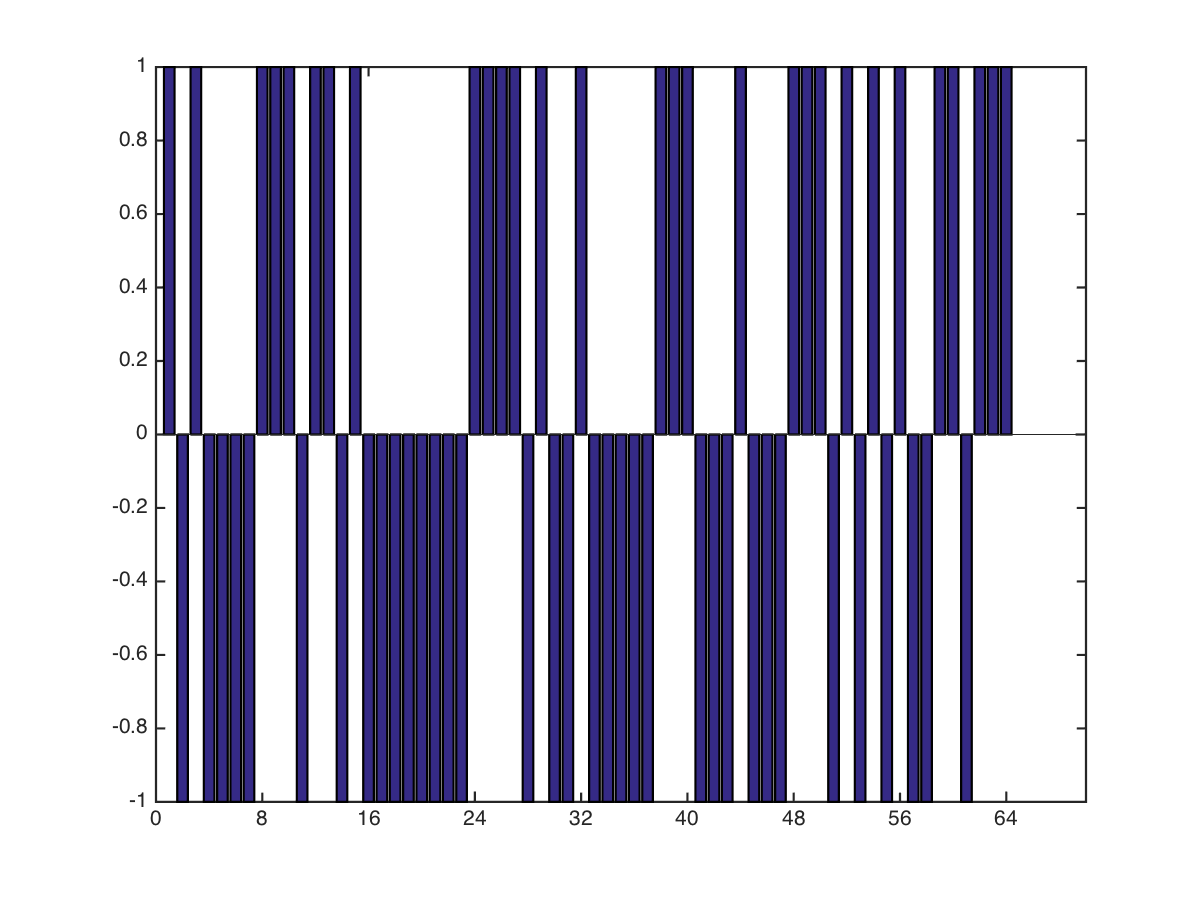}
}
\caption{The learned representations and binary hashing codes of one test image from CIFAR10, through: (a) (b) NNH; (c) (d) SNNH; (e) (f) proposed.}
\label{outputs}
\end{figure}

%on a workstation with 12 Intel Xeon 2.67GHz CPUs and 1 GTX680 GPU

\noindent \textbf{Datasets}
The \textbf{CIFAR10} dataset \cite{cifar} contains 60K labeled images of 10 different classes. The images are represented using 384-dimensional GIST descriptors \cite{gist}. Following the classical setting in \cite{masci2013sparse}, we used a training set of 200 images for each class, and a disjoint query set of 100 images per class. The remaining 59K images are treated as database.

\textbf{NUS-WIDE} \cite{nus} is a dataset containing 270K annotated images from Flickr. Every images is associated with one or more of the different 81 concepts, and is described using a 500-dimensional bag-of-features. In training and evaluation, we followed the protocol of \cite{liu2011hashing}: two images were considered as neighbors if they share at least one common concept (only 21 most frequent concepts are considered). We use 100K pairs of images for training, and a query set of 100 images per concept in testing.

\noindent \textbf{Comparison Methods}  We compare the proposed deep $\ell_\infty$ siamese network to six state-of-the-art hashing methods:
\begin{itemize}
\item four representative ``shallow'' hashing methods: kernelized supervised hashing (KSH) \cite{KSH}, anchor graph hashing (AGH) \cite{liu2011hashing} (we compare with its two alternative forms: AGH1 and AGH2; see the original paper), parameter-sensitive hashing (PSH) \cite{SSH}, and LDA Hash (LH) \cite{DH} \footnote{Most of the results are collected from the comparison experiments in \cite{masci2013sparse}, under the same settings.}. 
\item two latest ``deep'' hashing methods: neural-network hashing (NNH) \cite{NNHash}, and 
sparse neural-network hashing (SNNH) \cite{masci2013sparse}.
\end{itemize}
Comparing the two ``deep'' competitors to the deep $\ell_\infty$ siamese network, the only difference among the three is the type of encoder adopted in each's twin columns, as listed in Table \ref{encoder}.  We re-implement the encoder parts of NNH and SNNH, with three hidden layers (i.e, two unfolded stages for LISTA), so that all three deep hashing models have the same depth\footnote{The performance is thus improved than reported in their original papers using two hidden layers, although with extra complexity.}. Recall that the input $y \in R^n$ and the hash code $x \in R^N$, we immediately see from (\ref{recur}) that $\mathbf{W} \in R^{n \times N}$, $\mathbf{S}_k \in R^{N \times N}$, and $b_k \in R^N$. We carefully ensure that both NNHash and SparseHash have all their weight layers of the same dimensionality with ours\footnote{Both the deep $\ell_\infty$ encoder and the LISTA network will introduce the diagonal layers, while the generic feed-forward networks not. Besides, neither LISTA nor generic feed-forward networks contain layer-wise biases. Yet since either a diagonal layer or a bias contains only $N$ free parameters, the total amount is ignorable.}, for a fair comparison.

\begin{table*}[htbp]\small
\vspace{0mm}
\caption{Performance (\%)  of different hashing methods on the NUS-WIDE dataset, with different code lengths $N$. 
\vspace{0mm}
}
\begin{center}
\begin{tabular}{c c ccc  | c   c c c c c c c c}
& &&&		& &  & \multicolumn{3}{c}{{\bf Hamming radius} $\leq 2$} & \multicolumn{3}{c}{{\bf Hamming radius} $=0$} \\
\cline{8-13}
 \multicolumn{5}{c|}{{\bf Method}} 	& {\bf mAP@10}  & \hspace{-2mm}{\bf MP@5K}\hspace{-2mm} & {\bf Prec.} & {\bf Recall} & {\bf F1} & {\bf Prec.} & {\bf Recall} & {\bf F1} \\
\hline 
 &&  & & $N$
& & & & & & & & \\
\multirow{2}{*}{{\bf KSH}} 
& &   &&64			& 72.85 & 42.74 & 83.80 & 6.1$\times 10^{-3}$ & 1.2$\times 10^{-2}$ & 84.21 & 1.7$\times 10^{-3}$ & 3.3$\times 10^{-3}$\\
%& &  &&&80		& 72.76 & 43.32 & 84.21 & 1.8$\times 10^{-3}$ & 3.6$\times 10^{-3}$ & 84.23 & 1.4$\times 10^{-3}$ & 2.9$\times 10^{-3}$ \\
& & &&256			& 73.73 & 45.35 & 84.24 & 1.4$\times 10^{-3}$ & 2.9$\times 10^{-3}$ & 84.24 & 1.4$\times 10^{-3}$ & 2.9$\times 10^{-3}$ \\
\hline
\multirow{2}{*}{{\bf AGH1}} 
&&  &&64		& 69.48 & 47.28 & 69.43 & 0.11 & 0.22 & 73.35 & 3.9$\times 10^{-2}$ & 7.9$\times 10^{-2}$ \\
%&&  &&&80		& 69.62 & 47.23 & 71.15 & 7.5$\times 10^{-2}$ & 0.15 & 74.14 & 2.5$\times 10^{-3}$ & 5.1$\times 10^{-2}$ \\
&&  &&256	& 73.86 & 46.68 & 75.90 & 1.5$\times 10^{-2}$ & 2.9$\times 10^{-2}$ & 81.64 & 3.6$\times 10^{-3}$ & 7.1$\times 10^{-3}$ \\
\hline
\multirow{2}{*}{{\bf AGH2}} 
&& &&64		& 68.90 & 47.27 & 68.73 & 0.14 & 0.28 & 72.82 & 5.2$\times 10^{-2}$ & 0.10 \\
%&& &&&80	& 69.73 & 47.32 & 70.57 & 0.12 & 0.24 & 73.85 & 4.2$\times 10^{-2}$ & 8.3$\times 10^{-2}$ \\
&& &&256	& 73.00 & 47.65 & 74.90 & 5.3$\times 10^{-2}$ & 0.11 & 80.45 & 1.1$\times 10^{-2}$ & 2.2$\times 10^{-2}$ \\
\hline
\multirow{2}{*}{{\bf PSH}} 
&& &&64		& 72.17 & 44.79 & 60.06 & 0.12 & 0.24 & 81.73 & 1.1$\times 10^{-2}$ & 2.2$\times 10^{-2}$ \\
%&& &&&80			& 72.58 & 46.96 & 83.96 & 1.9$\times 10^{-3}$ & 3.9$\times 10^{-3}$ & 80.91 & 1.3$\times 10^{-2}$ & 2.6$\times 10^{-2}$ \\
&& &&256		& 73.52 & 47.13 & 84.18 & 1.8$\times 10^{-3}$ & 3.5$\times 10^{-3}$ & 84.24 & 1.5$\times 10^{-3}$ & 2.9$\times 10^{-3}$ \\
\hline
\multirow{2}{*}{{\bf LH}} 
&& &&64		& 71.33 & 41.69 & \textbf{84.26} & 1.4$\times 10^{-3}$ & 2.9$\times 10^{-3}$ & 84.24 & 1.4$\times 10^{-3}$ & 2.9$\times 10^{-3}$ \\
%&& &&&80	& 70.34 & 37.75 & 84.24 & 4.9$\times 10^{-3}$ & 9.8$\times 10^{-3}$ & 84.24 & 4.9$\times 10^{-3}$ & 9.8$\times 10^{-3}$ \\
&& &&256	& 70.73 & 39.02 & 84.24 & 1.4$\times 10^{-3}$ & 2.9$\times 10^{-3}$ & 84.24 & 1.4$\times 10^{-3}$ & 2.9$\times 10^{-3}$ \\
\hline
%
%& & &&$m$ & $M$ 
%& & & & & & & &  \\
%\cline{2-5}
%
\multirow{2}{*}{{\bf NNH}} & &&&\multirow{1}{*}{64}  
%& 4
%	& 76.52 & 57.37 & 78.94 & 4.1$\times 10^{-2}$ & 8.2$\times 10^{-2}$ & 83.92 & 3.0$\times 10^{-3}$ & 6.0$\times 10^{-3}$ \\
%&& &&  
%& 7
	& 76.39 & 59.76 & 75.51 & 1.59 & 3.11 & 81.24 & 0.10 & 0.20 \\
%	\cline{5-14}
%
%&&&&\multirow{1}{*}{80}  & 11
%	& 75.51 & 59.59 & 77.17 & 2.02 & 3.94  & 81.89 & 0.24 & 0.48 \\
%	\cline{5-14}
%
&&&&\multirow{1}{*}{256}  %& 4
%	& 77.70 & 59.96 & 84.23 & 1.6$\times 10^{-3}$ & 2.9$\times 10^{-3}$ & 84.24 & 1.4$\times 10^{-3}$ & 2.9$\times 10^{-3}$ \\
%&& &&  & 7
%	& 78.49 & 59.78 & 84.25 & 1.4$\times 10^{-3}$ & 2.9$\times 10^{-3}$ & 84.24 & 1.4$\times 10^{-3}$ & 2.9$\times 10^{-3}$ \\
%&& &&  
%& 16
	& 78.31 & 61.21 & 83.46 & 5.8$\times 10^{-2}$ & 0.11 & 83.94 & 4.9$\times 10^{-3}$ & 9.8$\times 10^{-3}$ \\
\hline
\multirow{2}{*}{{\bf SNNH}} & &&&\multirow{1}{*}{64}  
%& 4
%	& 76.52 & 57.37 & 78.94 & 4.1$\times 10^{-2}$ & 8.2$\times 10^{-2}$ & 83.92 & 3.0$\times 10^{-3}$ & 6.0$\times 10^{-3}$ \\
%&& &&  
%& 7
	& 74.87 & 56.82 & 72.32 & \textbf{1.99} & \textbf{3.87} & 81.98 & \textbf{0.37} & \textbf{0.73} \\
%	\cline{5-14}
%
%&&&&\multirow{1}{*}{80}  & 11
%	& 75.51 & 59.59 & 77.17 & 2.02 & 3.94  & 81.89 & 0.24 & 0.48 \\
%	\cline{5-14}
%
&&&&\multirow{1}{*}{256}  %& 4
%	& 77.70 & 59.96 & 84.23 & 1.6$\times 10^{-3}$ & 2.9$\times 10^{-3}$ & 84.24 & 1.4$\times 10^{-3}$ & 2.9$\times 10^{-3}$ \\
%&& &&  & 7
%	& 78.49 & 59.78 & 84.25 & 1.4$\times 10^{-3}$ & 2.9$\times 10^{-3}$ & 84.24 & 1.4$\times 10^{-3}$ & 2.9$\times 10^{-3}$ \\
%&& &&  
%& 16
	& 74.73 & 59.37 & 80.98 & \textbf{0.10} & \textbf{0.19} & 82.85 & \textbf{0.98}  & \textbf{1.94} \\
\hline
\multirow{2}{*}{{\bf Proposed}} & &&&\multirow{1}{*}{64}  
%& 4
%	& 76.52 & 57.37 & 78.94 & 4.1$\times 10^{-2}$ & 8.2$\times 10^{-2}$ & 83.92 & 3.0$\times 10^{-3}$ & 6.0$\times 10^{-3}$ \\
%&& &&  
%& 7
	& \textbf{79.89} & \textbf{63.04} & 79.95 & 1.72 & 3.38 & \textbf{86.23} & 0.30 & 0.60 \\
%	\cline{5-14}
%
%&&&&\multirow{1}{*}{80}  & 11
%	& 75.51 & 59.59 & 77.17 & 2.02 & 3.94  & 81.89 & 0.24 & 0.48 \\
%	\cline{5-14}
%
&&&&\multirow{1}{*}{256}  %& 4
%	& 77.70 & 59.96 & 84.23 & 1.6$\times 10^{-3}$ & 2.9$\times 10^{-3}$ & 84.24 & 1.4$\times 10^{-3}$ & 2.9$\times 10^{-3}$ \\
%&& &&  & 7
%	& 78.49 & 59.78 & 84.25 & 1.4$\times 10^{-3}$ & 2.9$\times 10^{-3}$ & 84.24 & 1.4$\times 10^{-3}$ & 2.9$\times 10^{-3}$ \\
%&& &&  
%& 16
	& \textbf{80.02} & \textbf{65.62} & \textbf{84.63} & 7.2$\times 10^{-2}$ & 0.15 & \textbf{89.49} & 0.57 & 1.13 \\
\hline
\end{tabular} 
\end{center}
\label{nus}\vspace{0mm}
\end{table*}%

We adopt the following classical criteria for evaluation: 1) \textit{precision and recall} (PR) for different Hamming radii, and the \textit{F1 score} as their harmonic average; 2) \textit{mean average precision} (MAP) \cite{muller2001performance}. Besides, for NUS-WIDE, as computing mAP is slow over this large dataset, we follow the convention of \cite{masci2013sparse} to compute the \textit{mean precision} (MP) of top-5K returned neighbors (MP@5K), as well as report mAP of top-10 results (mAP@10).

% \begin{figure}[htbp]
%\centering
%\begin{minipage}{0.30\textwidth}
%\centering \subfigure[] {
%\includegraphics[width=\textwidth]{precision.png}
%}\end{minipage}\\
%\begin{minipage}{0.30\textwidth}
%\centering \subfigure[] {
%\includegraphics[width=\textwidth]{recall.png}
%}\end{minipage}
%\caption{The comparison of three deep hashing methods on NUS-WIDE: (a) precision curve; (b) recall curve, both w.r.t the hashing code length $N$, within the Hamming radius of 2.}
%\label{curve}
%\end{figure}

 \begin{figure}[htbp]
\centering \subfigure[] {
\includegraphics[width=0.22\textwidth]{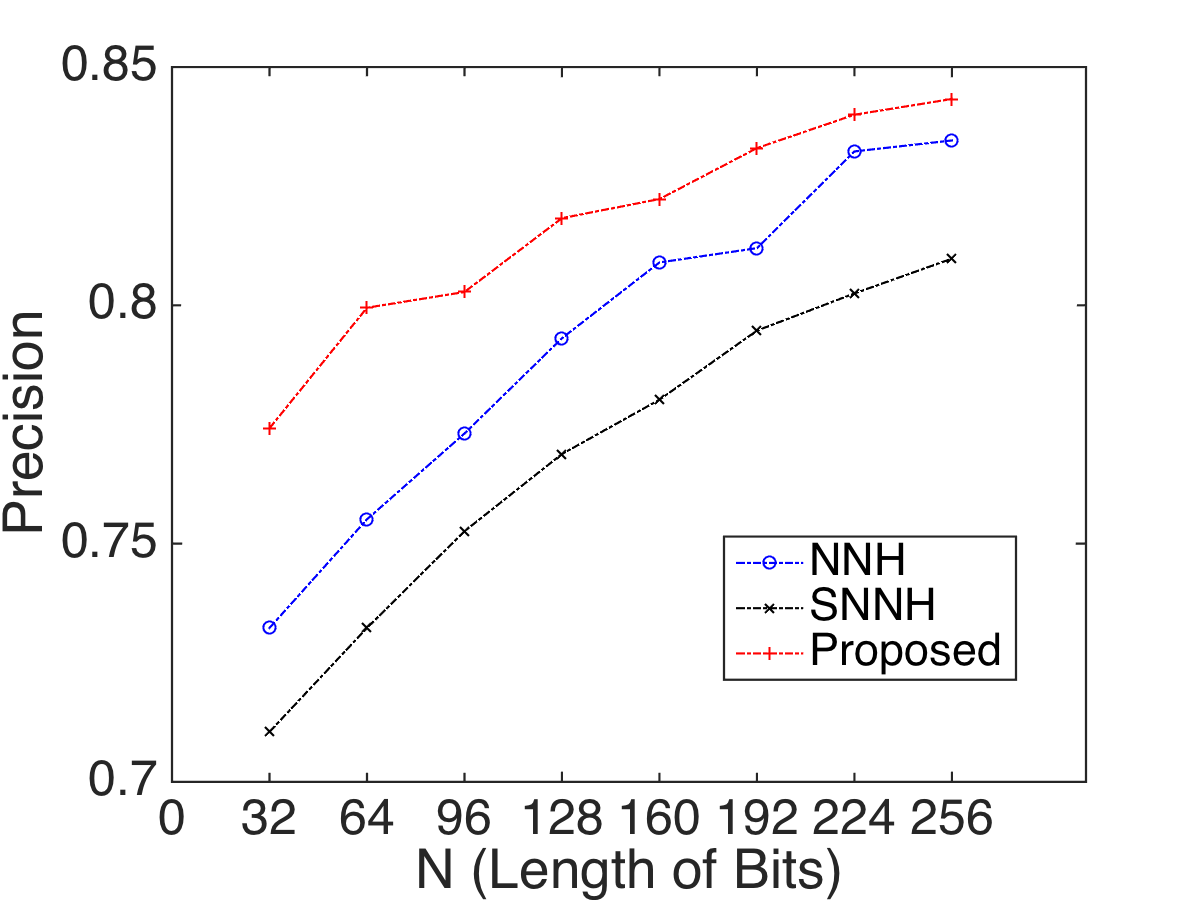}
}
\centering \subfigure[] {
\includegraphics[width=0.22\textwidth]{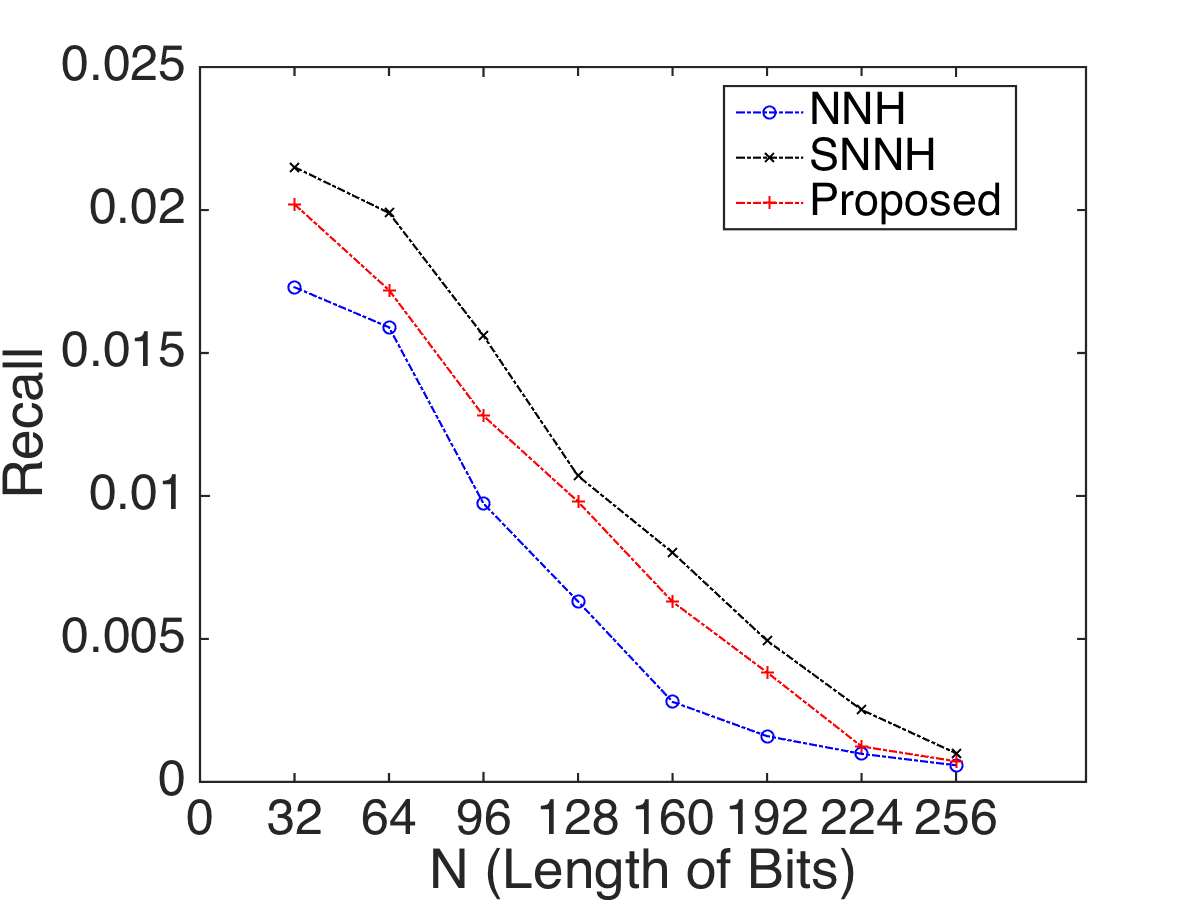}
}
\caption{The comparison of three deep hashing methods on NUS-WIDE: (a) precision curve; (b) recall curve, both w.r.t the hashing code length $N$, within the Hamming radius of 2.}
\label{curve}
\end{figure}

We have not compared with convolutional network-based hashing methods \cite{xia2014supervised,li2015feature,lai2015simultaneous}, since it is difficult to ensure their models to have the same parameter capacity as our fully-connected model in controlled experiments. We also do not include triplet loss-based methods, e.g., \cite{lai2015simultaneous}, into comparison because they will require three parallel encoder columns . 

%We plan to exploit convolutional layers as well as the triplet loss in our future work.

%\noindent \textbf{Evaluation Criteria} The following criteria are adopted: 
%\begin{itemize}
%\item \textit{precision and recall} (PR) for different Hamming radii, and the \textit{F1 score} as their
%harmonic average. 
%\item \textit{mean average precision}, defined as $mAP = \sum_{n} P(n)\cdot rel(n)$, where $rel(n)$ is the relevance of the $n$-th results (one if relevant and zero otherwise), and $P(n)$ is the precision at $n$ (i.e., percentage of relevant results in the  first $n$ top-ranked matches). 
%\item \textit{mean precision (MP)}, defined as the percentage of correct matches for a fixed number of retrieved elements. For NUS-WIDE, as computing mAP is slow over this large dataset, we show MP of top-5K returned neighbors (MP@5K), and mAP of top-10 results (mAP@10).
%\end{itemize}
% For the PR curves we use the ranking induced by the Hamming distance between the query and the database samples. In case of $r < m$ we considered only the results falling into the Hamming ball of radius $r$.

\noindent \textbf{Results and Analysis}   The performance of different methods on two datasets are compared in Tables \ref{cifar} and \ref{nus}. Our proposed method ranks top in almost all cases,  in terms of mAP/MP and precision. Even under the Hamming radius of 0, our precision result is as high as 33.30\% ($N$ = 64) for CIFAR10, and 89.49\% ($N$ = 256) for NUS-WIDE. The proposed method also maintains the second best in most cases, in terms of recall, inferior only to SNNH. In particular, when the hashing code dimensionality is low, e.g., when $N$ = 48 for CIFAR10, the proposed method outperforms all else with a significant margin. It demonstrates the competitiveness of the proposed method in generating both compact and accurate hashing codes, that achieves more precise retrieval results at lower computation and storage costs.

The next observation is that, compared to the strongest competitor SNNH, the recall rates of our method seem less compelling. We plot the precision and recall curves of the three best performers (NNH, SNNH, deep $l_\infty$), with regard to the bit length of hashing codes $N$, within the Hamming radius of 2. Fig. \ref{curve} demonstrates that our method consistently outperforms both SNNH and NNH in precision. On the other hand, SNNH gains advantages in recall over the proposed method, although the margin appears vanishing as $N$ grows.

Although it seems to be a reasonable performance tradeoff, we are curious about the behavior difference between SNNH and the proposed method. We are again reminded that they only differ in the encoder architecture, i.e., one with LISTA while the other using the deep $l_\infty$ encoder. We thus plot the learned representations and binary hashing codes of one CIFAR image, using NNH, SNNH, and the proposed method, in Fig. \ref{outputs}. By comparing the three pairs, one could see that the quantization from (a) to (b) (also (c) to (d)) suffer visible distortion and information loss. Contrary to them, the output of the deep $l_\infty$ encoder has a much smaller quantization error, as it  naturally resembles an antipodal signal. Therefore, it suffers minimal information loss during the quantization step. 

In view of those, we conclude the following points towards the different behaviors, between SNNH and deep $l_\infty$ encoder:
\begin{itemize}
\item Both deep $l_\infty$ encoder and SNNH outperform NNH, by introducing structure into the binary hashing codes.
\item The deep $l_\infty$ encoder generates nearly antipodal outputs that are robust to quantization errors. Therefore, it excels in preserving information against hierarchical information extraction as well as quantization. That explains why our method reaches the highest precisions, and performs especially well when $N$ is small.
\item SNNH exploits sparsity as a prior on hashing codes. It confines and shrinks the solution space, as many small entries in the SNNH outputs will be suppressed down to zero. That is also evidenced by Table 2 in \cite{masci2013sparse}, i.e., the number of unique hashing codes in SNNH results is one order smaller than that of NNH. 
\item The sparsity prior improves the recall rate, since its obtained hashing codes can be clustered more compactly in high-dimensional space, with lower intra-clutser variations. But it also runs the risk of losing too much information, during the hierarchical sparsifying process. In that case, the inter-cluster variations might also be compromised, which causes the decrease in precision. 
\end{itemize}
Further, it seems that the sparsity and $l_\infty$ structure priors could be complementary. We will explore it as future work.

\section{Conclusion}

This paper investigates how to import the quantization-robust property of an $\ell_\infty$-constrained minimization model, to a specially-designed deep model. It is done by first deriving an ADMM algorithm, which is then re-formulated as a feed-forward neural network. We introduce the siamese architecture concatenated with a pairwise loss, for the application purpose of hashing. We analyze in depth the performance and behaviors of the proposed model against its competitors, and hope it will evoke more interests from the community.

%\newpage

%% The file named.bst is a bibliography style file for BibTeX 0.99c
\bibliographystyle{named}
\bibliography{ijcai16}

\end{document}